\documentclass{IEEEtran}
\IEEEoverridecommandlockouts

\usepackage{graphicx}
\usepackage{subcaption}
\usepackage[style=base]{caption}
\usepackage{booktabs}

\usepackage{multirow}
\usepackage{amsmath,amssymb,amsfonts}
\usepackage{algorithmic}
\usepackage{graphicx}
\usepackage{textcomp}
\usepackage{xcolor}
\usepackage{cite}
\usepackage{hyperref}

\begin{document}

\title{Automatic Classification of Pathology Reports \\using TF-IDF Features}

\author{\IEEEauthorblockN{Shivam Kalra, Larry Li, H.R. Tizhoosh}\\
\IEEEauthorblockA{\textit{Kimia Lab, University of Waterloo, Canada, http://kimia@uwaterloo.ca}}
}

\maketitle

\begin{abstract} Pathology report is arguably one of the most important
  documents in medicine containing interpretive information about the visual
  findings from the patient's biopsy sample. Each pathology report has a
  retention period of up to 20 years after the treatment of a patient. Cancer
  registries process and encode high volumes of free-text pathology reports for
  surveillance of cancer and tumor diseases all across the world. In spite of
  their extremely valuable information they hold, pathology reports are not used
  in any systematic way to facilitate computational pathology.  Therefore, in
  this study we investigate automated machine-learning techniques to
  identify/predict the primary diagnosis (based on ICD-O code) from pathology
  reports. We performed experiments by extracting the TF-IDF features from the
  reports and classifying them using three different methods---SVM, XGBoost, and
  Logistic Regression. We constructed a new dataset with 1,949 pathology reports
  arranged into 37 ICD-O categories, collected from four different primary
  sites, namely lung, kidney, thymus, and testis. The reports were manually
  transcribed into text format after collecting them as PDF files from NCI
  Genomic Data Commons public dataset. We subsequently pre-processed the reports
  by removing irrelevant textual artefacts produced by OCR software. The highest
  classification accuracy we achieved was 92\% using XGBoost classifier on
  TF-IDF feature vectors, the linear SVM scored 87\% accuracy. Furthermore, the
  study shows that TF-IDF vectors are suitable for highlighting the important
  keywords within a report which can be helpful for the cancer research and
  diagnostic workflow. The results are encouraging in demonstrating the
  potential of machine learning methods for classification and encoding of
  pathology reports.
  
\end{abstract}
\section{Introduction}
Cancer is one of the leading causes of death in the world, with over 80,000
deaths registered in Canada in 2017 (Canadian Cancer Statistics 2017). A
computer-aided system for cancer diagnosis usually involves a pathologist
rendering a descriptive report after examining the tissue glass slides obtained
from the biopsy of a patient. A pathology report contains specific analysis of
cells and tissues, and other histopathological indicators that are crucial for
diagnosing malignancies. An average sized laboratory may produces a large
quantity of pathology reports annually (e.g., in excess of 50,000), but these
reports are written in mostly unstructured text and with no direct link to the
tissue sample. Furthermore, the report for each patient is a personalized
document and offers very high variability in terminology due to lack of
standards and may even include misspellings and missing punctuation, clinical
diagnoses interspersed with complex explanations, different terminology to label
the same malignancy, and information about multiple carcinoma appearances
included in a single report~\cite{GaoHierarchicalattentionnetworks2017}.

In Canada, each Provincial and Territorial Cancer Registry (PTCR) is responsible
for collecting the data about cancer diseases and reporting them to Statistics
Canada (StatCan). Every year, Canadian Cancer Registry (CCR) uses the
information sources of StatCan to compile an annual report on cancer and tumor
diseases. Many countries have their own cancer registry programs. These programs
rely on the acquisition of diagnostic, treatment, and outcome information
through manual processing and interpretation from various unstructured sources
(e.g., pathology reports, autopsy/laboratory reports, medical billing
summaries). The manual classification of cancer pathology reports is a
challenging, time-consuming task and requires extensive training
\cite{GaoHierarchicalattentionnetworks2017}.

With the continued growth in the number of cancer patients, and the increase in
treatment complexity, cancer registries face a significant challenge in manually
reviewing the large quantity of reports
\cite{weegar2017efficient,GaoHierarchicalattentionnetworks2017}. In this
situation, Natural Language Processing (NLP) systems can offer a unique
opportunity to automatically encode the unstructured reports into structured
data. Since, the registries already have access to the large quantity of
historically labeled and encoded reports, a supervised machine learning approach
of feature extraction and classification is a compelling direction for making
their workflow more effective and streamlined. If successful, such a solution
would enable processing reports in much lesser time allowing trained personnel
to focus on their research and analysis. However, developing an automated
solution with high accuracy and consistency across wide variety of reports is a
challenging problem.

For cancer registries, an important piece of information in a pathology report
is the associated ICD-O code which describes the patient's histological
diagnosis, as described by the World Health Organization's (WHO) International
Classification of Diseases for Oncology~\cite{louis20072007}. Prediction of the
primary diagnosis from a pathology report provides a valuable starting point for
exploration of machine learning techniques for automated cancer surveillance. A
major application for this purpose would be ``auto-reporting'' based on analysis
of whole slide images, the digitization of the biopsy glass slides. Structured,
summarized and categorized reports can be associated with the image content when
searching in large archives. Such as system would be able to drastically
increase the efficiency of diagnostic processes for the majority of cases where
in spite of obvious primary diagnosis, still time and effort is required from
the pathologists to write a descriptive report.

The primary objective of our study is to analyze the efficacy of existing
machine learning approaches for the automated classification of pathology
reports into different diagnosis categories. We demonstrate that TF-IDF feature
vectors combined with linear SVM or XGBoost classifier can be an effective
method for classification of the reports, achieving up to 83\% accuracy. We also
show that TF-IDF features are capable of identifying important keywords within a
pathology report. Furthermore, we have created a new dataset consisting of 1,949
pathology reports across 37 primary diagnoses. Taken together, our exploratory
experiments with a newly introduced dataset on pathology reports opens many new
opportunities for researchers to develop a scalable and automatic information
extraction from unstructured pathology reports.

\section{Background}
NLP approaches for information extraction within the biomedical research areas
range from rule-based systems~\cite{kang2012using}, to domain-specific systems
using feature-based classification~\cite{weegar2017efficient}, to the recent
deep networks for end-to-end feature extraction and
classification~\cite{GaoHierarchicalattentionnetworks2017}. NLP has had varied
degree of success with free-text pathology reports~\cite{wieneke2015validation}.
Various studies have acknowledge the success of NLP in interpreting pathology
reports, especially for classification tasks or extracting a single
attribute from a report~\cite{wieneke2015validation,imler2013natural}.

The Cancer Text Information Extraction System (caTIES)~\cite{crowley2010caties}
is a framework developed in a caBIG project focuses on information extraction
from pathology reports. Specifically, caTIES extracts information from surgical
pathology reports (SPR) with good precision as well as recall.

Another system known as Open Registry~\cite{contiero2008comparison} is capable
of filtering the reports with disease codes containing cancer.
In~\cite{d2010evaluation}, an approach called Automated Retrieval Console (ARC)
is proposed which uses machine learning models to predict the degree of
association of a given pathology or radiology with the cancer. The performance
ranges from an F-measure of 0.75 for lung cancer to 0.94 for colorectal cancer.
However, ARC uses domain-specific rules which hiders with the generalization of
the approach to variety of pathology reports.

This research work is inspired by themes emerging in many of the above studies.
Specifically, we are evaluating the task of predicting the primary diagnosis
from the pathology report. Unlike previous approaches, the system does not rely
on custom rule-based knowledge, domain specific features, balanced dataset with
fewer number of classes.

\section{Materials and Methods}
\begin{table}
  \caption{Distribution of pathology reports across~(a)~Primary diagnosis, used a the label for the study, and (b)~Primary site associated with a report.}
  \label{tab:report-distribution}
  \centering
  \begin{tabular}{lr}
    \toprule
    \multicolumn{2}{c}{(a) Primary Diagnosis}                                         \\
    \addlinespace[0.5ex]
    \midrule\addlinespace[0.7ex]
    Description                                          & Count                  \\
    \midrule
    Clear cell adenocarcinoma, NOS                       & 523                    \\
    Squamous cell carcinoma, NOS                         & 340                    \\
    Papillary adenocarcinoma, NOS                        & 300                    \\
    Adenocarcinoma, NOS                                  & 233                    \\
    Renal cell carcinoma, chromophobe type               & 113                    \\
    Adenocarcinoma with mixed subtypes                   & 89                     \\
    Seminoma, NOS                                        & 68                     \\
    Thymoma, type AB, malignant                          & 31                     \\
    Mixed germ cell tumor                                & 30                     \\
    Thymoma, type B2, malignant                          & 26                     \\
    Embryonal carcinoma, NOS                             & 26                     \\
    Thymoma, type A, malignant                           & 15                     \\
    Renal cell carcinoma, NOS                            & 14                     \\
    Thymoma, type B1, malignant                          & 13                     \\
    Bronchiolo-alveolar carcinoma, non-mucinous          & 13                     \\
    Thymoma, type B3, malignant                          & 13                     \\
    Acinar cell carcinoma                                & 13                     \\
    Mucinous adenocarcinoma                              & 11                     \\
    Thymic carcinoma, NOS                                & 11                     \\
    Basaloid squamous cell carcinoma                     & 9                      \\
    Thymoma, type AB, NOS                                & 7                      \\
    Squamous cell carcinoma, keratinizing, NOS           & 7                      \\
    Teratoma, benign                                     & 6                      \\
    Solid carcinoma, NOS                                 & 5                      \\
    Thymoma, type B2, NOS                                & 5                      \\
    Yolk sac tumor                                       & 4                      \\
    Papillary squamous cell carcinoma                    & 4                      \\
    Bronchiolo-alveolar adenocarcinoma, NOS              & 3                      \\
    Bronchio-alveolar carcinoma, mucinous                & 3                      \\
    Teratoma, malignant, NOS                             & 3                      \\
    Micropapillary carcinoma, NOS                        & 2                      \\
    Thymoma, type A, NOS                                 & 2                      \\
    Teratocarcinoma                                      & 2                      \\
    Squamous cell carcinoma, large cell, nonkeratinizing & 2                      \\
    Thymoma, type B1, NOS                                & 1                      \\
    Squamous cell carcinoma, small cell, nonkeratinizing & 1                      \\
    Signet ring cell carcinoma                           & 1                      \\    
    \midrule
    \multicolumn{2}{c}{(b) Primary Site} \\
    \midrule\addlinespace[0.2ex]
 
    Kidney                                               & 937                    \\
    Lung                                                 & 749                    \\
    Testis                                               & 139                    \\
    Thymus                                               & 124                    \\
    \bottomrule
  \end{tabular}
\end{table}


We assembled a dataset of 1,949 cleaned pathology reports. Each report is associated with
one of the 37 different primary diagnoses based on IDC-O codes. The reports are
collected from four different body parts or primary sites from multiple
patients. The distribution of reports across different primary diagnoses and
primary sites is reported in~\autoref{tab:report-distribution}. The dataset was
developed in three steps as follows.

\textbf{Collecting pathology reports:} The total of 11,112 pathology reports
were downloaded from NCI's Genomic Data Commons (GDC) dataset in PDF
format~\cite{grossman2016toward}. Out of all PDF files, 1,949 reports were
selected across multiple patients from four specific primary sites---thymus,
testis, lung, and kidney. The selection was primarily made based on the quality
of PDF files.

\textbf{Cleaning reports:} The next step was to extract the text
content from these reports. Due to the significant time expense of manually re-typing all the
pathology reports, we developed a new strategy to prepare our dataset. We
applied an Optical Character Recognition (OCR) software to convert the PDF
reports to text files. Then, we manually inspected all generated text files
to fix any grammar/spelling issues and irrelevant characters as an artefact produced by the OCR system.

\textbf{Splitting into training-testing data: } We split the cleaned reports
into 70\% and 30\% for training and testing, respectively. This split resulted
in 1,364 training, and 585 testing reports.

\subsection{Pre-Processing of Reports}
We pre-processed the reports by setting their text content to lowercase and
filtering out any non-alphanumeric characters. We used NLTK library to remove
stopping words, e.g., `the', `an', `was', `if' and so on \cite{LoperNLTKNaturalLanguage2002}. We then analyzed the reports to find
common bigrams, such as ``lung parenchyma'', ``microscopic examination'', ``lymph node''
etc. We joined the biagrams with a hyphen, converting them into a single word.
We further removed the words that occur less than 2\% in each of the diagnostic
category. As well, we removed the words that occur more than 90\% across all the
categories. We stored each pre-processed report in a separate text file.

\subsection{TF-IDF features}
TF-IDF stands for \emph{Term Frequency-Inverse Document Frequency}, and it is a useful
weighting scheme in information retrieval and text mining. TF-IDF signifies the
importance of a term in a document within a corpus. It is important to note
that a \emph{document} here refers to a pathology report, a \emph{corpus} refers to the collection of
reports, and a \emph{term} refers to a single word in a report. The TF-IDF weight
for a term $t$ in a document $d$ is given by
\begin{align}
  \begin{split}
    TF(t, d) &= \frac{\text{Number of times $t$ appears in $d$}}{\text{Total number terms in $d$}},\\
    IDF(t) &= log\Big(\frac{\text{Total number of documents}}{\text{Number of documents with $t$}}\Big),\\
    TF\text{-}IDF(t, d) &= TF(t, d) * IDF(t)\\
  \end{split}
\end{align}

We performed the following steps to transform a pathology report into a feature
vector:
\begin{enumerate}
\item Create a set of vocabulary containing all unique words from all the
  pre-processed training reports.
\item Create a zero vector $f_r$ of the same length as the vocabulary.
\item For each word $t$ in a report $r$, set the corresponding index in $f_r$ to
  $TF-IDF(t, r)$.
  \item The resultant $f_r$ is a feature vector for the report $r$ and it is a
    highly sparse vector.
  \end{enumerate}

\subsection{Keyword extraction and topic modelling}
The keyword extraction involves identifying important words within reports that summarizes its content. In contrast, the topic modelling allows grouping these keywords using an intelligent scheme, enabling users to further focus on certain aspects of a document. All the words in a pathology report are sorted according to their TF-IDF weights. The
top $n$ sorted words constitute the top $n$ keywords for the report. The $n$
is empirically set to 50 within this research. The extracted keywords are further grouped into
different topics by using latent Dirichlet allocation
(LDA)~\cite{blei2003latent}. The keywords in a report are highlighted using the color theme based on their topics.

\subsection{Evaluation metrics}
Each model is evaluated using two standard NLP metrics---micro and macro
averaged F-scores, the harmonic mean of related metrics precision and recall.
For each diagnostic category $C_j$ from a set of 37 different classes $C$,
the number of true positives $TP_j$, false positives $FP_j$, and false negatives
$FN_j$, the micro F-score is defined as
\begin{align} \label{eq:boom}
  \begin{split}
    P^{micro} &= \frac{\sum_{C_j}^{C}TP_j}{\sum_{C_j}^{C}(TP_j + FP_j)}\\
    R^{micro} &= \frac{\sum_{C_j}^{C}TP_j}{\sum_{C_j}^{C}(TP_j + FN_j)}\\
    F^{micro} &= \frac{2P^{micro}R^{micro}}{P^{micro}+R^{micro}},
  \end{split}
\end{align}
whereas macro F-score is given by
\begin{equation}
  \label{eq:1}
  F^{macro} = \frac{1}{|C|}\sum_{C_j}^{C}F(C_j).
\end{equation}

In summary, micro-averaged metrics have class representation roughly proportional
to their test set representation (same as accuracy for classification problem
with a single label per data point), whereas macro-averaged metrics are averaged
by class without weighting by class
prevalence~\cite{QiuDeepLearningAutomated2018}.

\subsection{Experimental setting}
In this study, we performed two different series of experiments: i) evaluating the
performance of TF-IDF features and various machine learning classifiers on the
task of predicting primary diagnosis from the text content of a given report,
and ii) using TF-IDF and LDA techniques to highlight the important keywords
within a report. For the first experiment series, training reports are
pre-processed, then their TF-IDF features are extracted. The TF-IDF features and
the training labels are used to train different classification models. These
different classification models and their hyper-parameters are reported
in \autoref{tab:classifier}. The performance of classifiers is measured
quantitatively on the test dataset using the evaluation metrics discussed in the
previous section. For the second experiment series, a random report is selected
and its top 50 keywords are extracted using TF-IDF weights. These 50 keywords
are highlighted using different colors based on their associated topic, which
are extracted through LDA. A non-expert based qualitative inspection is
performed on the extracted keywords and their corresponding topics.

\section{Results and Discussion}
\subsection{Experiment Series 1}
A classification model is trained to predict the primary diagnosis given the
content of the cancer pathology report. The performance results on this task are
reported in \autoref{tab:results}. We can observe that the XGBoost classifier
outperformed all other models for both the micro F-score metric, with a score of
0.92, and the macro F-score metric, with a score of 0.31. This was an
improvement of 7\% for the micro F-score over the next best model, SVM-L, and a
marginal improvement of 5\% for macro F-score. It is interesting to note that
SVM with linear kernels performs much better than SVM with RBF kernel, scoring
9\% on the macro F-score and 12\% more on the micro F-score. It is suspected
that since words used in primary diagnosis itself occur in some reports, thus enabling the linear
models to outperform complex models.

\begin{table}[ht]
  \caption{Different classifiers used in the study}
  \label{tab:classifier}
  \centering
  \begin{tabular}{lll}
    \toprule
    Code    & Classifier          & Parameters                             \\
    \midrule
    SVM-L   & SVM                 & kernel linear, C 1.0, shrinking true   \\
    SVM-RBF & SVM                 & kernel rbf, C 1.0, shrinking: true     \\
    LR      & Logistic Regression & penalty l2, solver liblinear, C 1.0 \\
    XGBoost & XGBoost             & max depth 6, learning rate 0.3         \\
    \bottomrule
  \end{tabular}
\end{table}

\begin{table}[ht]
  \caption{Final train and test performance of classification models}
  \label{tab:results}
  \centering
  \begin{tabular}{lccccc}
    \toprule
    \multirow{ 2}{*}{Classifier Code} & \multicolumn{2}{c}{Micro F-score} & \multicolumn{2}{c}{Macro F-score}             \\ \cmidrule(lr){2-3}\cmidrule(lr){4-5}
                                      & Train                             & Test          & Train         & Test          \\
    \midrule
    SVM-L                             & 0.95                              & 0.87          & 0.28          & 0.24          \\
    SVM-RBM                           & 0.80                              & 0.75          & 0.19          & 0.18          \\
    LR                                & 0.82                              & 0.78          & 0.20          & 0.18          \\
    XGBoost                           & \textbf{0.99}                     & \textbf{0.92} & \textbf{0.64} & \textbf{0.31} \\
    \bottomrule
  \end{tabular}
\end{table}


\subsection{Experiment Series 2}
\begin{figure}[htbp]
  \centering
  \includegraphics[width=0.95\linewidth]{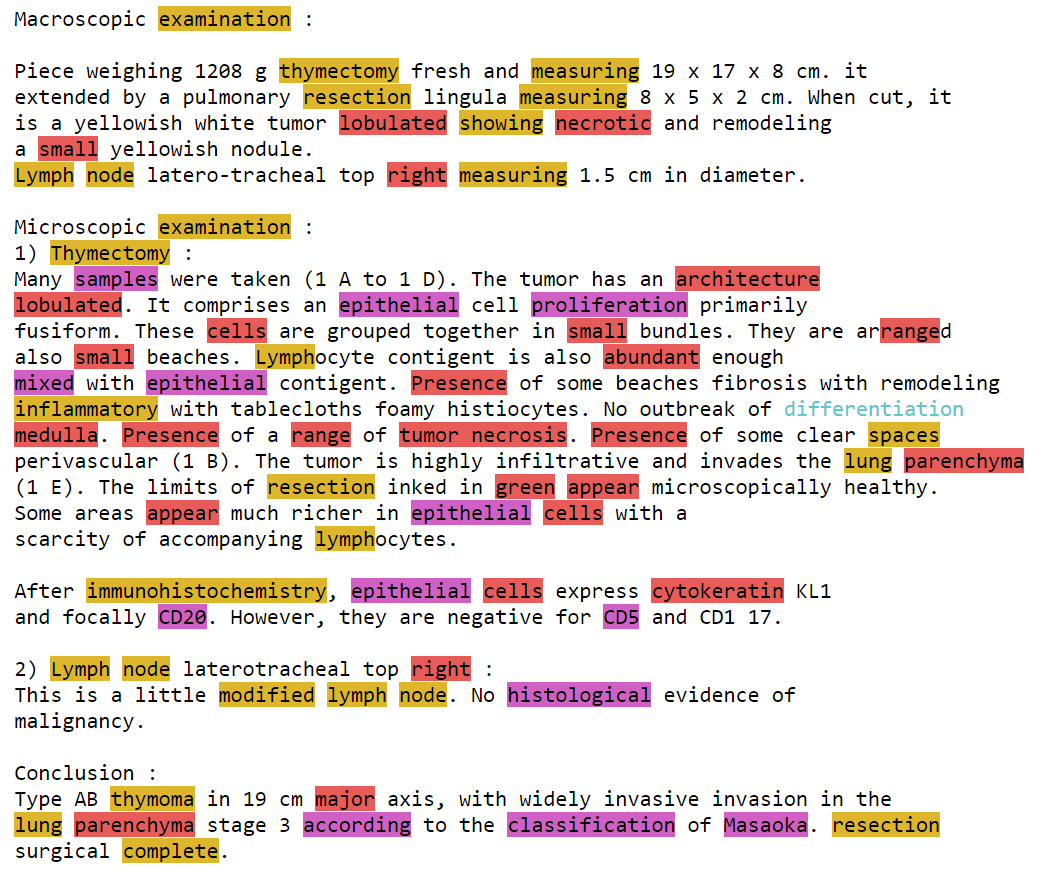}
  \scalebox{0.7}{
    \begin{tabular}{ll}
      \toprule
      \multicolumn{2}{c}{Top 10 Keywords}                                 \\
      \midrule
      \multicolumn{2}{l}{1. Epithelial (0.377), 2. Presence (0.269), 3. Thymectomy (0.232)} \\
      \multicolumn{2}{l}{4. Epithelial cells (0.210), 5. Cells (0.180), 6. Small (0.161), 7. Lobulated (0.161)} \\
      \multicolumn{2}{l}{8. Lung parenchyma (0.151),  9. Appear (0.150), 10. Examination (0.139)}   \\
      \midrule
      Topic \#                                          & \multicolumn{1}{c}{Keywords}                                      \\
      \midrule
      \multirow{3}{*}{\colorbox[HTML]{DCB32D}{Topic 1}} & examination, thymectomy, measuring, resection, showing,           \\
                                                        & inflammatory, lymph, spaces, lung, immunohistochemistry,          \\
                                                        & node, modified, complete                                          \\
      \multirow{2}{*}{\colorbox[HTML]{8B2E7F}{Topic 2}} & samples, epithelial, proliferation, mixed, CD20, CD5,             \\
                                                        & histological, according, classification, masaoka                  \\
      \multirow{3}{*}{\colorbox[HTML]{BD4443}{Topic 3}} & lobulated, necrotic, small, right, architecture,                  \\
                                                        & presence, medulla, range, tumor, necrosis, green, appear,         \\
                                                        & parenchyma, cells, cytokeratin, right, major                      \\
      \bottomrule      
    \end{tabular}
  }
  \caption{The top 50 keywords in a report identified using TF-IDF weights. The
    keywords are color encoded as per the
    abstract ``topics'' extracted using LDA. Each topic is given a separate
    color scheme.}
  \label{fig:keywords}
\end{figure}

\autoref{fig:keywords} shows the top 50 keywords highlighted using TF-IDF
and LDA. The proposed approach has performed well in highlighting the
important regions, for example the topic highlighted with a red color containing ``presence range tumor necrosis'' provides useful biomarker information to readers.

\subsection{Conclusions}
We proposed a simple yet efficient TF-IDF method to extract and corroborate useful keywords from pathology cancer reports. Encoding a pathology report for cancer and tumor surveillance is a
laborious task, and sometimes it is subjected to human errors and variability in the
interpretation. One of the most important aspects of encoding a
pathology report involves extracting the primary diagnosis. This may be very useful for content-based image retrieval to combine with visual information. We used existing classification model and TF-IDF features to predict the
primary diagnosis. We achieved up to 92\% accuracy using XGBoost classifier. The prediction accuracy empowers the adoption of machine learning methods for automated
information extraction from pathology reports.

\bibliography{ref}{}
\bibliographystyle{ieeetr}

\end{document}